\documentclass[10pt,twocolumn,letterpaper]{article}

\usepackage{cvpr}              

\newcommand{\ours}{parallel rescaling}
\newcommand{\oursp}{parallel rescaled}
\newcommand{\oursc}{Parallel Rescaling}

\definecolor{cvprblue}{rgb}{0.21,0.49,0.74}
\usepackage[pagebackref,breaklinks,colorlinks,allcolors=cvprblue]{hyperref}
\usepackage{algorithm}\usepackage{algorithmic}

\def\paperID{2} 
\def\confName{CVPR}
\def\confYear{2025}

\title{\oursc : Rebalancing Consistency Guidance for Personalized Diffusion Models}

\author{
JungWoo Chae\thanks{Equal contribution.} \\
Nexon Korea \\
\texttt{cjwnexon@nexon.co.kr}
\and
Jiyoon Kim\footnotemark[1] \\
LGCNS AI Research \\
\texttt{jiyoonkim@lgcns.com}
\and
Sangheum Hwang\thanks{Corresponding author} \\
Department of Data Science, \\
Seoul National University of Science and Technology \\
\texttt{shwang@seoultech.ac.kr}
}

\begin{document}
\maketitle
\begin{abstract}
Personalizing diffusion models to specific users or concepts remains challenging, particularly when only a few reference images are available. Existing methods such as DreamBooth and Textual Inversion often overfit to limited data, causing \textbf{misalignment between generated images and text prompts} when attempting to balance identity fidelity with prompt adherence. While Direct Consistency Optimization (DCO) with its consistency-guided sampling partially alleviates this issue, it still struggles with complex or stylized prompts. In this paper, we propose a \textbf{\ours \ technique} for personalized diffusion models. Our approach explicitly decomposes the consistency guidance signal into parallel and orthogonal components relative to classifier-free guidance (CFG). By rescaling the parallel component, we minimize disruptive interference with CFG while preserving the subject’s identity. Unlike prior personalization methods, our technique \textbf{does not} require additional training data or expensive annotations. Extensive experiments show improved prompt alignment and visual fidelity compared to baseline methods, even on challenging stylized prompts. These findings highlight the potential of \oursp~guidance to yield more stable and accurate personalization for diverse user inputs.
\end{abstract}

\section{Introduction}
\label{sec:intro}

Text-to-image diffusion models~\cite{sdxl,sd3} have transformed content creation by enabling users to generate vivid, imaginative visuals simply from textual prompts. Recently, personalization techniques such as DreamBooth~\cite{dreambooth} and Textual Inversion~\cite{ti} have expanded these capabilities further, allowing a user to incorporate a custom subject---e.g., a specific person, animal, or product---into generated images after fine-tuning with only a few examples. However, these personalized models often suffer from \emph{text misalignment}: they overfit to the limited training images, sometimes ignoring or overriding aspects of the prompt and unintentionally recreating backgrounds or styles from the reference set.

A notable effort to address overfitting is \emph{Direct Consistency Optimization (DCO)}~\cite{dco}, which introduces a consistency function to constrain the fine-tuned diffusion model so that its outputs remain close to those of the base (pre-trained) generator. Although DCO effectively balances \emph{identity fidelity} with \emph{prompt adherence}, it can still fail to perform reliably on \emph{long or complex prompts}, as well as highly \emph{stylized} descriptions. Meanwhile, other approaches typically require extensive resources, such as costly segmentation masks~\cite{break_a_scene,mudi} or high-VRAM hardware~\cite{attndreambooth, disendiff}, making them less accessible to the average user.

In this work, we propose a \emph{simple yet effective} method for personalization that preserves subject identity \emph{and} offers stronger alignment with complex prompts. Our key insight is to analyze the \emph{consistency guidance} signal by decomposing it into parallel and orthogonal components with respect to \emph{classifier-free guidance (CFG)} \cite{cfg}. Through this decomposition, we discover that the parallel component, which is essential for maintaining the target subject’s features, can inadvertently \emph{conflict} with text guidance---especially during denoising steps where stylization plays a critical role. Based on these observations, we introduce a \emph{\ours \ strategy} that mitigates interference in CFG-based guidance, thereby improving prompt fidelity while retaining subject identity. Our contributions are:
\begin{itemize}
    \item We perform a detailed analysis of the consistency guidance signal, showing how decomposing it into parallel and orthogonal components reveals the source of text misalignment in personalized diffusion.
    \item We propose a straightforward \emph{\ours} technique that re-centers and re-scales the parallel component of consistency guidance. This ensures stronger alignment with complex prompts while retaining subject features.
\end{itemize}

Through experiments, we demonstrate that our approach reliably preserves the subject’s identity while improving text alignment, even under elaborate or artistic prompts. By requiring only minimal computational overhead, our \ours~guidance represents a promising step toward more accessible and robust personalized diffusion models.

\section{Related Work}
\label{sec:rel_work}

\paragraph{Diffusion Models and Personalization.}
Recent diffusion models~\cite{sdxl,dalle3,sd3} have proven highly effective for text-to-image generation, creating diverse and high-fidelity images. Although these models exhibit impressive creative range, direct application to \emph{personalized} content remains challenging. Two leading approaches for personalization are \emph{DreamBooth}~\cite{dreambooth} and \emph{Textual Inversion}~\cite{ti}.  Building on these methods, recent work~\cite{customdiffusion, dreammatcher,p+, palp, neti,oft} has further expanded the field with techniques such as subject-driven generation~\cite{wei2023elite}, identity-preserving diffusion~\cite{ma2023subject}, and segmentation-based personalization~\cite{break_a_scene,mudi}. Both traditional and newer techniques highlight an ongoing tension between \emph{identity fidelity} and \emph{prompt fidelity} in personalized diffusion.

\vspace{-10pt}
\paragraph{Guidance Methods.} To steer text-to-image diffusion models during sampling, \emph{Classifier-Free Guidance (CFG)}~\cite{cfg} scales the difference between conditional and unconditional predictions, boosting alignment with the user's prompt. While CFG often improves adherence, excessive scaling can distort visual quality and reduce diversity. Various enhancements address these limitations: for instance, \emph{autoguidance}~\cite{autoguidance} adaptively tunes the guidance scale to prevent mode collapse, \emph{guidance interval}~\cite{guidanceinterval} selectively applies CFG at specific timesteps for smoother sampling. More recent approaches like  \emph{Attend-and-Excite}~\cite{attend} further refine guidance by leveraging attention mechanisms. However, these strategies primarily target generic text-to-image tasks and do not inherently resolve the \emph{personalization} trade-off between subject preservation and stylistic or prompt-based variation.

\emph{Direct Consistency Optimization (DCO)}~\cite{dco} specifically tackles this trade-off by learning a consistency function that anchors a fine-tuned model’s outputs to those of a base diffusion model. During inference, a \emph{Consistency Guidance} term is added to CFG to preserve the subject’s core features. However, a fixed consistency weight can still interfere with the prompt—particularly under complex or stylized conditions. Our work builds on DCO’s insights by decomposing the consistency guidance into parallel and orthogonal components relative to CFG. Through a \emph{\ours \  strategy}, we mitigate disruptive interactions in the parallel term, enabling finer control of prompt fidelity with minimal sacrifice of identity preservation. In doing so, we contribute to the ongoing effort to balance personalization with the versatility needed to handle complex prompts.
\section{Preliminary: Consistency Guidance}

\textbf{Classifier-Free Guidance (CFG).} In diffusion models, CFG steers the generation toward a given text prompt without an external classifier. At each denoising step $t$, the model is run twice: once with conditional prompt $c$ and once without prompt (the ``unconditional'' case). Let $\epsilon_\phi(x_t|c)$ denote the noise prediction conditioned on $c$ using the pretrained model (with parameters $\phi$) and $\epsilon_\phi(x_t|\varnothing)$ denote the prediction for an empty prompt. The text guidance vector is defined as:
\begin{equation}
g_{\text{text}}(x_t) = \epsilon_\phi(x_t|c) - \epsilon_\phi(x_t|\varnothing),
\end{equation}
and is scaled by a factor \( \omega_{\text{text}} > 1 \) (the guidance scale) to form the guided prediction:
\begin{equation}
\epsilon_{\text{CFG}}(x_t) = \epsilon_\phi(x_t|\varnothing) + \omega_{\text{text}}\,g_{\text{text}}(x_t).
\end{equation}

\noindent\textbf{Consistency Guidance in DCO.} To enforce consistency with reference images, Consistency Guidance Sampling extends CFG by incorporating an additional guidance term. After fine-tuning the model to learn a \emph{consistency function} that measures similarity between generated images and reference images, DCO defines a consistency condition \( c_{\text{cons}} \) that anchors the model to the base concept. The consistency guidance vector is given by:
\begin{equation}
g_{\text{cons}}(x_t) = \epsilon_\theta(x_t\,|\,c) - \epsilon_\phi(x_t|\,c),
\end{equation}
where \( \epsilon_\theta \) represents the noise prediction from the personalized (fine-tuned) model. The final sampler update combines both text and consistency guidance:
\begin{equation}
\epsilon_{\text{CG}}(x_t) = \epsilon_\phi(x_t|\varnothing) + \omega_{\text{text}}\,g_{\text{text}}(x_t) + \omega_{\text{cons}}\,g_{\text{cons}}(x_t),
\end{equation}
where \( \omega_{\text{cons}} \) controls the influence of the consistency term. The magnitudes of \( \omega_{\text{text}} \) and \( \omega_{\text{cons}} \) can be adjusted according to user preference to control the trade-off between adherence to the text prompt and consistency with the reference images.
\section{Parallel Rescaling of Consistency Guidance}
\label{sec:method}

We propose a \ours \ strategy for the consistency guidance term~\cite{dco} in personalized diffusion. Our approach is to decompose and re-scale the portion of $g_{\text{cons}}$ that aligns with the text guidance $g_{\text{text}}$, because an excessively large parallel component can diminish or override prompt details.

\subsection{Decomposition and Motivation}
\label{subsec:decomposition}

Let $g_{\text{cons}}$ be the consistency guidance vector that preserves a subject’s features during sampling. We split it into parallel and orthogonal parts with respect to the text guidance $g_{\text{text}}$  \emph{per location} $(w,h)$:
\begin{equation}
\label{eq:decompose}
    g_{\text{cons}}
    \;=\;
    g_{\text{cons}}^{\parallel}
    \;+\;
    g_{\text{cons}}^{\perp}.
\end{equation}
Here, $g_{\text{cons}}^{\parallel}$ projects onto $g_{\text{text}}$, meaning it can \emph{reinforce} or \emph{interfere} with the prompt, while $g_{\text{cons}}^{\perp}$ retains the subject’s identity in directions unrelated to the text.  
\\[3pt]
\noindent
\textbf{Why decompose along $g_{\text{text}}$?}  
If the parallel component of $g_{\text{cons}}$ becomes overly large, it can overshadow the text guidance signal and degrade prompt fidelity. By isolating $g_{\text{cons}}^{\parallel}$, we can selectively re-scale it without discarding the beneficial identity information in $g_{\text{cons}}^{\perp}$.

\subsection{Measuring Interference: \texorpdfstring{\textbf{$\text{Consistency}_p$}}{Consistency p}}
\label{subsec:cp}

To understand how $g_{\text{cons}}^{\parallel}$ interacts with $g_{\text{text}}$, we define:
\begin{equation}
\label{eq:consistency_p}
    \text{Consistency}_p(w,h)
    \;=\;
    \mathrm{mean}_{\mathrm{channel}}
    \Bigl(
        \frac{\omega_{\text{cons}}
              \cdot
              g_{\text{cons}}^{\parallel}(w,h)}
             {\omega_{\text{text}}
              \cdot
              g_{\text{text}}(w,h)}
    \Bigr).
\end{equation}
Here, $(w,h)$ denotes a location in the latent (or image) space, and $\mathrm{mean}_{\mathrm{channel}}(\cdot)$ averages across channels. This ratio indicates how strongly the parallel consistency term \emph{reinforces} ($> 0$) or \emph{opposes} ($< 0$) the text guidance. 

\begin{figure}[t]
\centering
\vspace{3pt}
\includegraphics[width=0.9\linewidth]{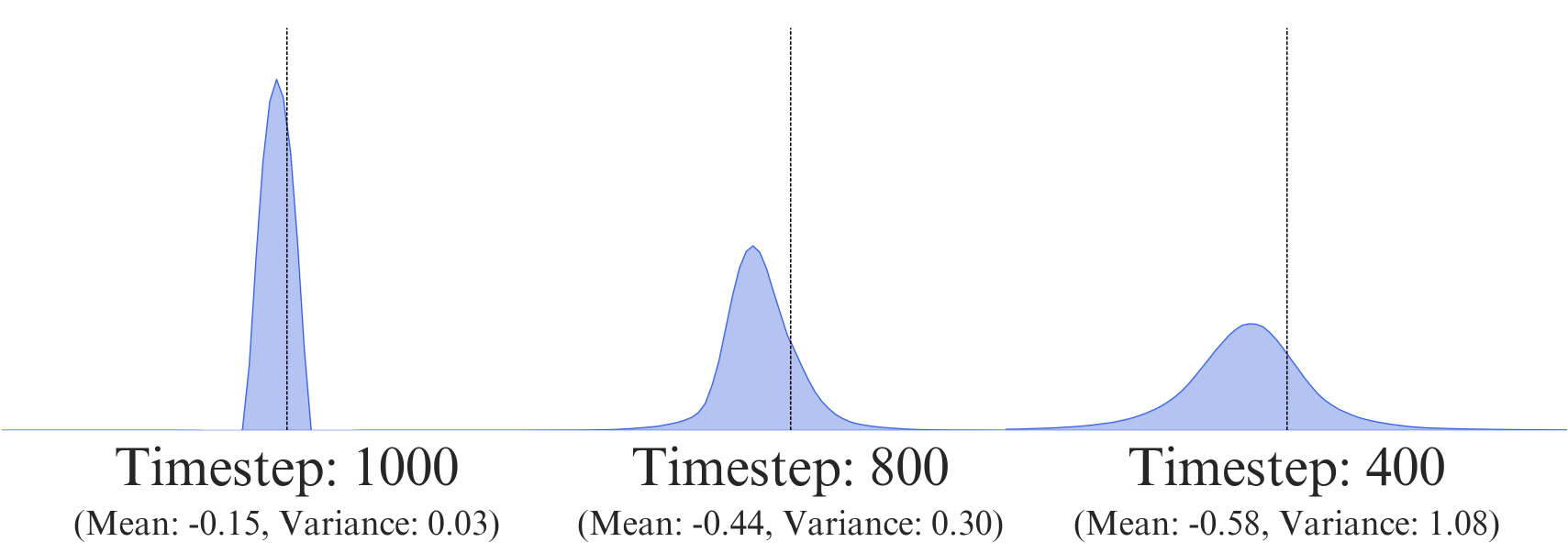}
\caption{
\textbf{Distribution shift of $\text{Consistency}_p$ as timesteps decrease.} 
The parallel $g_{cons}$ component drifts negatively, counteracting the text guidance for stylized prompts. 
}
\label{fig:distribution_shift}
\vspace{-1em}
\end{figure}

\vspace{1pt}
\noindent
\textbf{Distribution Shift.}  
As denoising proceeds, $\text{Consistency}_p$ tends to skew negative and grow in variance (see Fig.~\ref{fig:distribution_shift}). Large negative values reduce $g_{\text{text}}$’s influence, thereby weakening stylization or complex scene details. In other words, an excessively negative parallel component inevitably undermines the prompt direction.

\subsection{Parallel Rescaling}
\label{subsec:parallel_rescaling}

To suppress the directional shift and stabilize the variance of $\text{Consistency}_p$, we re-center and re-scale the parallel signal by :
\vspace{-2em}

{\small
\begin{equation}
\label{eq:normalization}
\begin{aligned}
    g_{\text{PR}}
    =\;&
    g_{\text{cons}}^{\perp}+
    \frac{\omega_{\text{text}}}{\omega_{\text{cons}}} \cdot
    \frac{\text{Consistency}_p - \mu(\text{Consistency}_p)}
         {\sigma(\text{Consistency}_p) + \epsilon}
    \odot g_{\text{text}},
\end{aligned}
\end{equation}
}
where $\mu(\cdot)$ and $\sigma(\cdot)$ denote the mean and standard deviation of $\text{Consistency}_p$ over all spatial locations, and $\epsilon$ is a small constant (e.g., $3\times10^{-8}$). Note that the second term on the RHS represents the rescaled parallel component by our definition of $\text{Consistency}_p$ in Eq.~\ref{eq:consistency_p}.
\\[3pt]
\noindent
\textbf{Interpretation.}  
By normalizing and re-scaling the parallel term, we control how strongly $g_{\text{cons}}$ interferes with $g_{\text{text}}$. This ensures that the subject’s features are preserved \emph{without} excessively weakening prompt-based stylization.

\subsection{Sampling Procedure}
\label{subsec:sampling}

Algorithm~\ref{alg:parallel_rescaling} outlines our sampling steps. At each diffusion timestep $t$, we (i) decompose $g_{\text{cons}}$ into parallel and orthogonal parts; (ii) compute and rescale $\text{Consistency}_p$; and (iii) update $g_{\text{PR}}$ according to Eq.~\eqref{eq:normalization}. 

\begin{algorithm}[t]
\small
\caption{Parallel Rescaling of Consistency Guidance}
\label{alg:parallel_rescaling}
\begin{algorithmic}[1]
\REQUIRE 
    Personalized model $\theta$,
    Base model $\phi$,
    Prompt $c$,
    Guidance scales $\omega_{\text{text}}, \omega_{\text{cons}}$
\STATE Sample $x_T$ from $\mathcal{N}(0,I)$ 
\FOR{$t = T$ to $1$}
    \STATE $g_{\text{text}} \leftarrow \epsilon_\phi(x_t \mid c) - \epsilon_\phi(x_t \mid \varnothing)$
    \STATE $g_{\text{cons}} \leftarrow \epsilon_\theta(x_t \mid c) - \epsilon_\phi(x_t \mid c)$
    \STATE Decompose $g_{\text{cons}}$ into
    $g_{\text{cons}}^{\parallel}$ and $g_{\text{cons}}^{\perp}$ 
    \STATE Compute $\text{Consistency}_p$ from Eq.~\eqref{eq:consistency_p}, then re-scale it (Eq.~\ref{eq:normalization})
    \STATE $\epsilon_{\text{final}} \leftarrow
      \epsilon_\phi(x_t \mid \varnothing) 
      + \omega_{\text{text}}\,g_{\text{text}}
      + \omega_{\text{cons}}\,g_{\text{PR}}$
    \STATE Apply diffusion update on $x_t$ with $\epsilon_{\text{final}}$
\ENDFOR
\RETURN $x_0$
\end{algorithmic}
\end{algorithm}

\section{Experiments}
\label{sec:experiments}

This section describes our experimental setup, the evaluation metrics, and both quantitative and qualitative comparisons of our proposed method against multiple baselines.

\subsection{Setup}
We use Stable Diffusion XL (SDXL)~\cite{sdxl} as our base model. For personalization, we consider two approaches: DreamBooth + Textual Inversion (TI)~\cite{dreambooth} and Direct Consistency Optimization (DCO)~\cite{dco}, each fine-tuned with LoRA (rank=32) for efficiency. We compare:
\begin{itemize}
    \item \textbf{CFG}: Standard classifier-free guidance without consistency.
    \item \textbf{Consistency Guidance (CG)}: Baseline method adding a consistency-derived term~\cite{dco}.
    \item \textbf{Ours (\ours)}: Our proposed \ours~of the parallel consistency component.
\end{itemize}
We evaluate on 16 categories (e.g., objects, plush toys, animals) commonly used for personalization, setting \(\omega_{\text{text}}=7.5\) and \(\omega_{\text{cons}}=3.0\) unless otherwise noted. For quantitative evaluation, we employ 20 distinct prompts and generate 4 images per prompt for each method, yielding 80 samples per personalization approach.

\begin{figure*}[t]
\centering
\includegraphics[width=\linewidth]{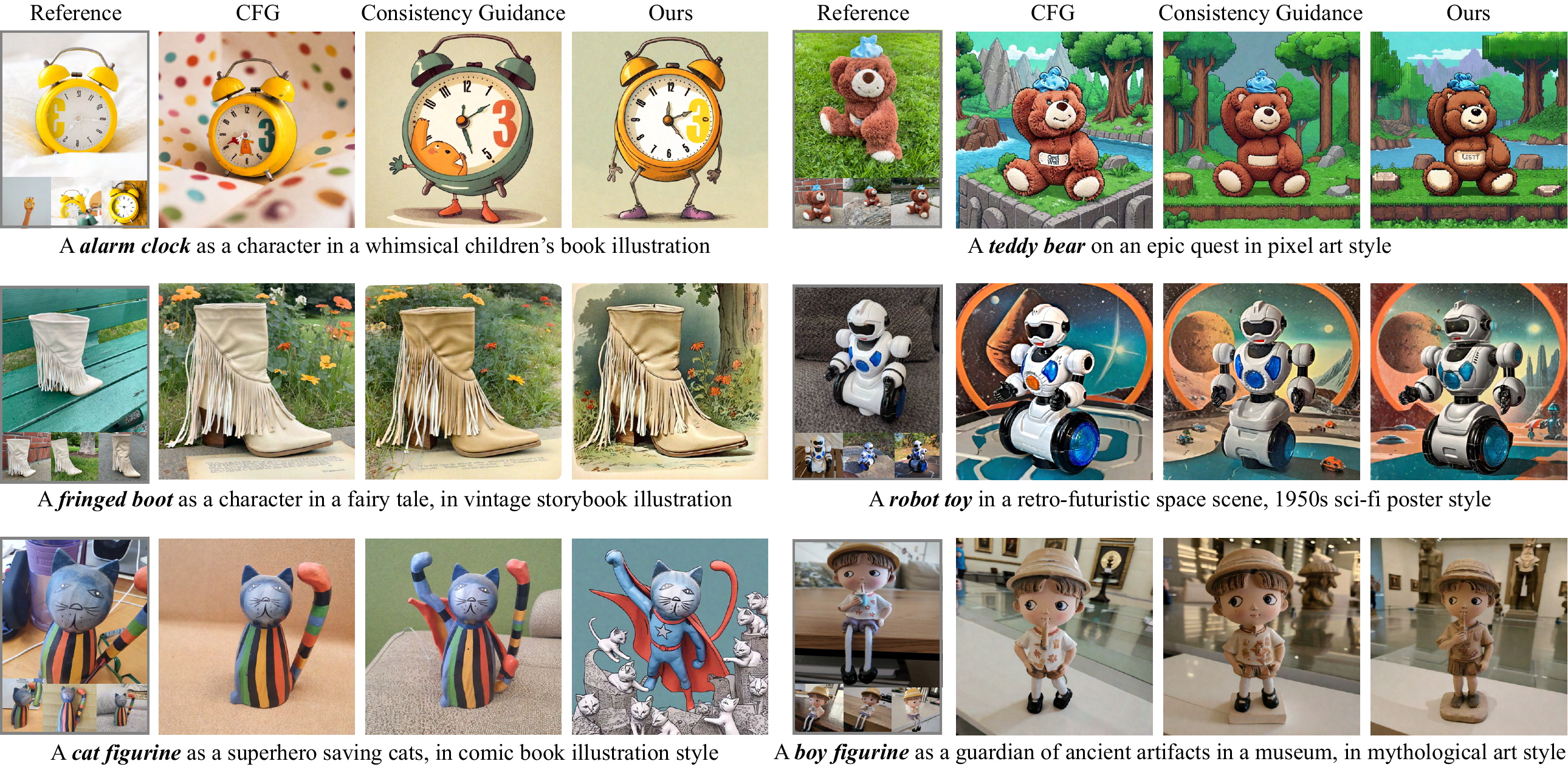}
\vspace{-8mm}
\caption{Qualitative results: Examples show our method preserving subject identity and stylization across a variety of creative prompts.}
\label{fig:qualitative}
\vspace{-5mm}
\end{figure*}

\subsection{Evaluation Metrics}
We assess two primary metrics:
\begin{itemize}
    \item \textbf{Text-Image Alignment:} A CLIP-based~\cite{CLIP} similarity between the generated images and their input prompts.
    \item \textbf{Identity Preservation:} An image-similarity score via DINOv2~\cite{DINOv2}, comparing each generated image to the reference set for that concept.
\end{itemize}
Additionally, we prepare \emph{challenging prompts} that emphasize stylization (artistic or cartoon-like rendering) and complex subject-text interactions to test the robustness of each method.

\subsection{Quantitative Results}
\noindent In Table~\ref{tab:quantitative}, we present results for models fine-tuned via DreamBooth+TI, comparing a baseline Consistency Guidance (CG) sampler to our \ours. The baseline obtains moderate text alignment and higher identity fidelity, yet it often struggles with stylized or intricate prompts. Our method, while incurring a minor trade-off in image similarity, demonstrates stronger text alignment. This improved \emph{prompt adherence} is especially valuable for stylization scenarios.

\begin{table}[t]
\centering
\begin{tabular}{l|l|cc}
\hline
\textbf{Training} & \textbf{Method}  & \textbf{Text Align.} & \textbf{Image Sim.} \\
\hline
{DB+TI} & CFG & 0.6383 & 0.7033 \\
        & CG & 0.6457 & 0.6833 \\
       & Ours & 0.6517 & 0.6776 \\
\hline
DCO &  CFG & 0.6391 & 0.7047 \\
    & CG         & 0.6482 & 0.6790 \\
    & Ours       & 0.6534 & 0.6765 \\
\hline
\end{tabular}
\vspace{-2mm}
\caption{Partial Quantitative Results. We report mean CLIP-based text alignment and DINO-based image similarity. ``DB+TI'' indicates DreamBooth + Textual Inversion, while ``DCO'' indicates Direct Consistency Optimization.} 
\label{tab:quantitative}
\vspace{-5mm}
\end{table}

\subsection{Qualitative Results}
\label{subsec:qual}
\noindent 
We perform qualitative comparisons on prompts demanding both \emph{stylization} and \emph{subject-text coherence}.
Under such conditions, Consistency Guidance (CG) often struggles, occasionally reverting to photorealistic traits or simpler backgrounds and thus failing to reflect stylized details.
In contrast, our method (\ours) consistently enforces more faithful stylization while retaining subject identity.
Figure~\ref{fig:qualitative} shows examples where \ours~preserves the subject's features under a wide range of creative prompt styles.
Additional examples illustrating a variety of artistic prompts are provided in Fig.~\ref{fig:extra_qual_a} in the Appendix.
\section{Conclusion}
\label{sec:conclusion}

We propose a parallel rescaling method that minimizes the identity-fidelity/prompt-adherence trade-off in personalized diffusion by selectively re-centering and re-scaling the parallel component of consistency guidance to mitigate interference with classifier-free guidance. Experiments show notable gains in prompt adherence and visual quality, especially for stylized or complex prompts, without needing additional training data or annotations. 

\vspace{3pt}
\noindent
\textbf{Limitations and Future Work.}
While effective, this straightforward normalization strategy may not fully resolve all conflicts between text alignment and subject identity. Future work could explore more adaptive or prompt-aware weighting schemes for further refinement.

{
    \small
    \bibliographystyle{ieeenat_fullname}
    \bibliography{main}
}

\clearpage
\clearpage
\setcounter{page}{1}
\setcounter{section}{0}
\maketitlesupplementary
\renewcommand{\thesection}{\Alph{section}}

\section{Implementation Details}
\label{appendix:implementation}

\noindent \textbf{Prompts for Image Generation.} 
We focus on \emph{complex, stylized prompts} to thoroughly test each model’s ability to preserve subject identity under challenging conditions. A selection of these prompts is given in Table~\ref{tab:prompt_list}, which spans both columns for readability.

\begin{table*}[t]
\centering
\caption{Complex Prompts Used in Our Experiments. Each prompt is crafted to test stylization, environmental details, and subject-text interplay. ``[V]'' denotes the placeholder token for the personalized subject.}
\label{tab:prompt_list}
\begin{tabular}{p{0.95\textwidth}}
\hline
\textbf{Prompts} \\
\hline
1. A [V] is building a sandcastle on a sunny beach while tiny crabs scuttle around and seagulls fly overhead.\\[3pt]
2. A [V] is lifting a barbel at the gym.\\[3pt]
3. A [V] wearing a police cap, resting on the police car.\\[3pt]
4. A photo of [V] made out of lego building blocks.\\[3pt]
5. A [V] surfing giant waves at sunset.\\[3pt]
6. A [V] dressed as a cowboy, riding a white fluffy donkey in the desert.\\[3pt]
7. A [V] as a Jedi casting a long shadow in a sunlit, empty desert.\\[3pt]
8. A [V] as navy officer, saluting at a naval parade with a crowd cheering, in a pastel drawing style.\\[3pt]
9. A [V] sprinting on a running track, painted in impressionist style.\\[3pt]
10. A [V] collecting nuts in an autumn forest, illustrated in art nouveau style.\\[3pt]
11. A painting of a [V] floating on the lake under the full moon's glow in the style of Monet.\\[3pt]
12. A [V] in a dramatic action scene in retro comic book.\\[3pt]
13. A [V] on an epic quest in pixel art style.\\[3pt]
14. [V], crashed down in distance Anime drawing, on mars.\\[3pt]
15. [V] riding a bicycle through a city park, urban sketch style.\\[3pt]
16. An illustration of [V], playing fetch with its owner in a serene meadow at dawn, in vintage poster style.\\[3pt]
17. A product overview page of [V] in the magazine, illustrated in a infographic style.\\[3pt]
18. A surreal painting of [V] in Magritte style.\\[3pt]
19. A [V] playing guitar in pop art style.\\[3pt]
20. An oil painting of a [V] dressed as a musketeer in an old French town.\\
\hline
\end{tabular}
\end{table*}

\vspace{0.5em}
\noindent \textbf{Personalization Training Configuration.}
\begin{itemize}
    \item Base Model: Stable Diffusion XL (SDXL).
    \item Personalization Approaches: 
    \begin{itemize}
        \item \emph{DreamBooth + Textual Inversion (TI)}~\cite{dreambooth}
        \item \emph{Direct Consistency Optimization (DCO)}~\cite{dco}, with $\beta=1000$
    \end{itemize}
    \item LoRA Setup: Low-rank adaptation of rank 32.
    \item Hyperparameters:
    \begin{itemize}
        \item \emph{Batch Size:} 1 image per iteration
        \item \emph{Training Steps:} 1000 steps
        \item \emph{Learning Rates:} 5e-5
        \item \emph{Text Encoder Learning Rates:} 5e-6
        \item \emph{Optimizer:} AdamW ($\beta_1=0.9$, $\beta_2=0.999$)
    \end{itemize}
    \item Hardware: All experiments conducted on an NVIDIA RTX 3090 GPU with 24 GB VRAM (or similar).
\end{itemize}

\vspace{0.5em}
\noindent \textbf{Inference Configuration.}
\begin{itemize}
    \item \emph{Sampling Steps:} 50 DDIM steps
    \item \emph{Guidance Scales:} $\omega_{\text{text}}=7.5$, $\omega_{\text{cons}}=3.0$ (for methods using consistency)
\end{itemize}

\section{Additional Qualitative Visualizations}
\label{appendix:visuals}

\begin{figure*}[ht]
    \centering
    \includegraphics[width=\linewidth]{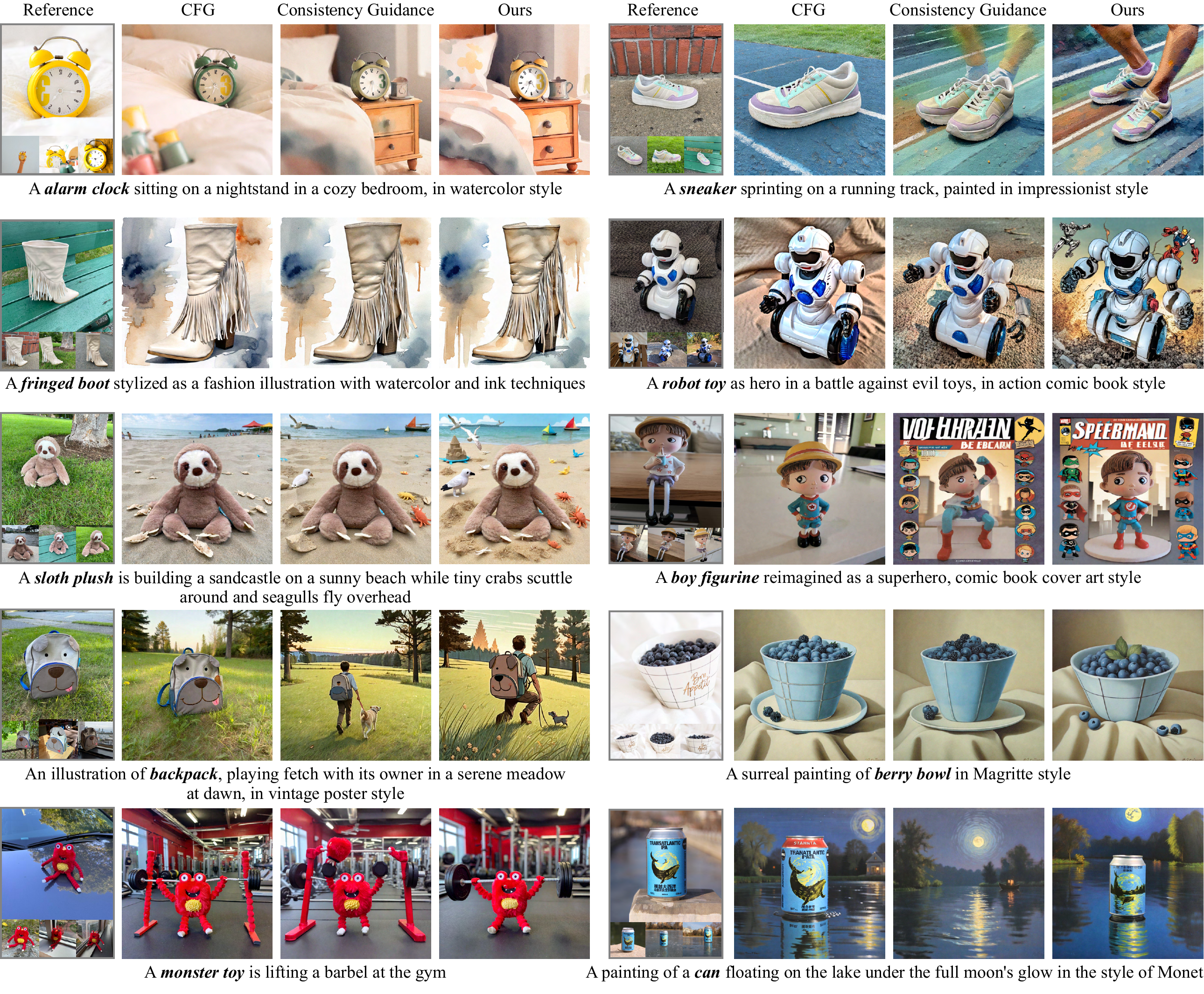}
    \caption{Additional qualitative comparison under stylized prompts. From left to right: Reference images, CFG, Consistency Guidance, Ours (\ours).}
    \label{fig:extra_qual_a}
\end{figure*}

Figures~\ref{fig:extra_qual_a} show additional results comparing Consistency Guidance versus our Parallel Rescaling Guidance, across diverse prompts and subject identities. These examples further highlight the ability of our approach to handle stylized requests without sacrificing identity fidelity or prompt coherence.

\newpage
\clearpage  
\setcounter{page}{9}


\end{document}